\documentclass[runningheads]{llncs}
\usepackage[T1]{fontenc}
\usepackage{bbm}  
\usepackage{amsmath,amssymb}
\usepackage{booktabs}
\usepackage{multirow}
\usepackage{graphicx,verbatim}
\begin{document}
\title{BrainSTR: Spatio-Temporal Contrastive Learning for Interpretable Dynamic Brain Network Modeling}
\titlerunning{BrainSTR for Interpretable Dynamic Brain Network Modeling}

\author{
Guiliang Guo\inst{1} \and
Guangqi Wen\inst{2} \and
Lingwen Liu\inst{3} \and
Ruoxian Song\inst{1} \and
Peng Cao\inst{1}\thanks{Corresponding author.} \and
Jinzhu Yang\inst{1} \and
Fei Wang\inst{4} \and
Xiaoli Liu\inst{5} \and
Osmar R. Zaiane\inst{6}
}

\authorrunning{Guo et al.}

\institute{
Computer Science and Engineering, Northeastern University, Shenyang, China
\email{caopeng@cse.neu.edu.cn}
\and
School of Computer Science and Artificial Intelligence, Shandong Normal University, Jinan, China
\and
Graduate School of Information, Production and Systems, Waseda University, Kitakyushu, Fukuoka, Japan
\and
Nanjing Medical University, Nanjing, China
\and
AiShiWeiLai AI Research, Beijing, China
\and
Amii, University of Alberta, Edmonton, Alberta, Canada
}

\maketitle              % typeset the header of the contribution

\begin{abstract} Dynamic functional connectivity captures time-varying brain states for better neuropsychiatric diagnosis and {spatio-temporal interpretability}, i.e., identifying \emph{when} discriminative disease signatures emerge and \emph{where} they reside in the connectivity topology. 
Reliable interpretability faces major challenges: diagnostic signals are often subtle and sparsely distributed across both time and topology, while nuisance fluctuations and non-diagnostic connectivities are pervasive. To address these issues, we propose \textbf{BrainSTR}, a spatio-temporal contrastive learning framework for interpretable dynamic brain network modeling. 
BrainSTR learns state-consistent phase boundaries via a data-driven \emph{Adaptive Phase Partition} module, identifies diagnostically critical phases with attention, and extracts disease-related connectivity within each phase using an \emph{Incremental Graph Structure Generator} regularized by binarization, temporal smoothness, and sparsity.
Then, we introduce a spatio-temporal supervised contrastive learning approach that leverages diagnosis-relevant spatio-temporal patterns to refine the similarity metric between samples and capture more discriminative spatio-temporal features, thereby constructing a well-structured semantic space for coherent and interpretable representations. Experiments on ASD, BD, and MDD validate the effectiveness of BrainSTR, and the discovered critical phases and subnetworks provide interpretable evidence consistent with prior neuroimaging findings. Our code: \url{https://anonymous.4open.science/r/BrainSTR1}.

% Authors must provide keywords and are not allowed to remove this Keyword section.

\end{abstract}
\section{Introduction}
\label{sec:introduction}
Functional brain network analysis using resting-state fMRI (rs-fMRI) has shown great potential for assisting the diagnosis of neuropsychiatric disorders such as ASD, BD, and MDD~\cite{wang2022multi,wen2024heterogeneous}.
To characterize time-varying brain states, dynamic functional connectivity (dFC) is often estimated by constructing a sequence of time-resolved networks from BOLD signals, commonly using fixed-length sliding-window schemes or adaptive temporal partitioning strategies~\cite{liu2023braintgl,vsverko2022dynamic,jiang2025new}.
Despite its ability to model dynamic brain states, a key challenge in applying dFC to neuropsychiatric diagnosis is achieving spatio-temporal interpretability, i.e., pinpointing precisely when and where in the connectivity topology discriminative disease signatures emerge, to uncover reliable biomarkers.
Reliable spatio-temporal interpretability from dynamic brain networks remains challenging because disease-related signals are often subtle and sparsely distributed across both time and topology, whereas disease-irrelevant fluctuations and connectivities are pervasive.  
%\textbf{Topologically}, phase-specific (i.e., based on temporally contiguous brain-state segments)  Pearson correlation (PCC) graphs are often dense and contaminated with disease-irrelevant or noisy connectivities, producing spurious patterns~\cite{yoshikawa2020heart,zhang2023brainusl,heinsfeld2018identification}.
%\textbf{Temporally}, brain dynamics involve many latent phases, many of which capture transient or physiological fluctuations and are low- or non-discriminative~\cite{iraji2021tools,li2024exploring}.
% To address these issues, we propose a spatio-temporal contrastive learning framework, termed BrainSTR, for spatio-temporal interpretability in dynamic brain networks. In contrast to conventional contrastive learning, which suffers from interference by excessive disease-irrelevant information in brain networks, thereby compromising its ability to construct a precise similarity metric. It inevitably results in a lower similarity among subjects with potentially similar (positive pairs). 
% We aim to construct a well-structured semantic space by jointly selecting \textcolor{red}{temporally critical phases and identifying topologically disease-relevant connectivity patterns.}

To address these issues, we propose BrainSTR, a spatio-temporal contrastive learning framework for interpretable dynamic brain network modeling. 
Conventional contrastive learning suffers from interference by excessive disease-irrelevant information in brain networks, which compromises similarity estimation and leads to reduced similarity even among semantically similar subjects (positive pairs). 
In contrast, BrainSTR constructs a well-structured semantic space by jointly selecting temporally critical phases and identifying topologically disease-relevant connectivity patterns.
Specifically, we learn state-consistent phase boundaries via a data-driven Adaptive Phase Partition (APP) module and identify a compact set of diagnostically critical phases with attention.
Meanwhile, we propose an incremental graph structure generator, which is regularized with binarization, temporal smoothness, and sparsity terms to yield stable and interpretable connectivity selection for retaining disease-related connectivities while filtering out disease-unrelated topology within each phase. With the spatio-temporal disease-related representations, we apply supervised contrastive learning to encourage an optimal spatio-temporal semantic embedding space where the intrinsic similarity patterns of the spatio-temporal data are captured. 
Finally, this overall spatio-temporal interpretable representation learning is jointly optimized through critical-phase identification and incremental graph structure generator to construct a disease-relevant embedding space.
% Experiments on ASD, BD, and MDD show that BrainSTR consistently outperforms strong state-of-the-art baselines across all three disorders, demonstrating robust cross-task generalization and clear performance margins. Moreover, the discovered critical phases and subnetworks provide interpretable evidence consistent with prior neuroimaging findings.
% Experiments on ASD, BD, and MDD show that BrainSTR achieves the best ACC/AUC on all three tasks, reaching 77.2/77.8 for MDD, 78.2/79.6 for BD, and 72.4/73.0 for ASD (NYU). 
% Compared with the strongest prior results, BrainSTR improves performance by +2.9/+6.5 points on MDD, +4.5/+4.9 on BD, and +3.0/+3.3 on ASD (ACC/AUC), respectively. 
Experiments on ASD, BD, and MDD show that BrainSTR achieves the best ACC/AUC on all three tasks, reaching 77.2\%/77.8\% for MDD, 78.2\%/79.6\% for BD, and 72.4\%/73.0\% for ASD (NYU).
Compared with the strongest prior results, BrainSTR improves performance by +2.9\%/+6.5\% on MDD, +4.5\%/+4.9\% on BD, and +3.0\%/+3.3\% on ASD (ACC/AUC), respectively.
These consistent gains indicate robust cross-disorder generalization with clear performance margins. 
Moreover, the discovered critical phases and subnetworks provide interpretable evidence consistent with prior neuroimaging findings.
\textbf{Our main contributions are:}
\begin{enumerate}
    % \item We propose \textbf{BrainSTR}, a contrastive learning framework that advances dynamic brain network diagnosis by shifting the focus from redundant whole-sample representations to the sparse yet critical disease-relevant spatio-temporal features.   
    \item We propose \textbf{BrainSTR}, a contrastive learning framework for dynamic brain network diagnosis that shifts the focus from redundant whole-sample representations to the sparse yet critical disease-relevant spatio-temporal features.
    \item We introduce a critical \textbf{spatio-temporal learning} scheme, involving adaptive phase partition for capturing diagnostically informative phases and incremental graph structure generator for learning disease-relevant connectivity patterns, reducing temporal noise and topological redundancy to improve robustness and interpretability.
    \item Extensive experiments on ASD, BD, and MDD demonstrate that BrainSTR \textbf{consistently outperforms state-of-the-art methods}, while ablation studies verify the necessity of each component. Moreover, BrainSTR discovers \textbf{clinically meaningful critical phases} and brain subnetworks that are consistent with prior neuroimaging findings, providing strong spatio-temporal interpretability.
\end{enumerate}

\begin{figure}
\includegraphics[width=\textwidth]{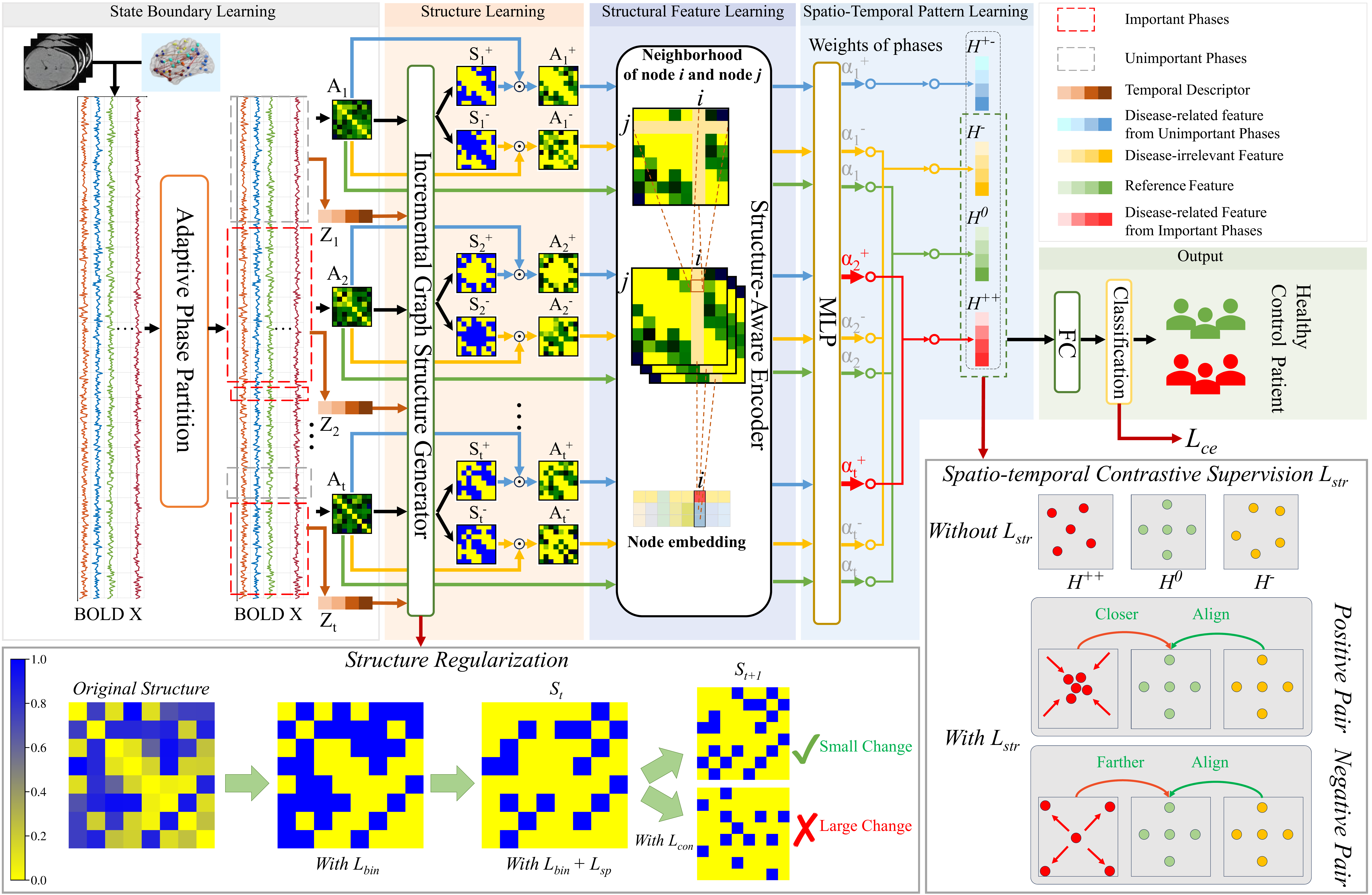}

\caption{Overview of the proposed \textbf{BrainSTR} framework.
Given a subject's BOLD signal $X\in\mathbb{R}^{T\times N}$, where $T$ and $N$ denote the numbers of time points and ROIs, respectively, Adaptive Phase Partition (APP) infers \emph{state-consistent} phase boundaries to construct phase-wise FCs $\{A_t\}_{t=1}^{W}$, where $W$ is the number of partitioned phases, with each $A_t\in\mathbb{R}^{N\times N}$.
% Each $A_t$ is decomposed into $A_t^{+}$/$A_t^{-}$ via a Graph Structure Generator and encoded by a shared Structure-Aware Encoder.
% Each $A_t$ is decomposed into $A_t^{+}$/$A_t^{-}$ via a Graph Structure Generator that learns a phase-wise structure $S_t$, and encoded by a shared Structure-Aware Encoder.
Each $A_t$ is decomposed into $A_t^{+}$/$A_t^{-}$ via an Incremental Graph Structure Generator that learns a phase-wise structure $S_t$, and then encoded by a shared Structure-Aware Encoder.
Attention then weights informative phases to form subject embeddings, while contrastive supervision makes them diagnosis-discriminative relative to original-graph embeddings.
}

\label{method}
\end{figure}

\section{Method}
\label{sec:Methods}

% Given a subject's BOLD signal $X\in\mathbb{R}^{T\times N}$, where $T$ and $N$ denote the number of time points and ROIs, respectively, APP infers \textcolor{red}{state-consistent} phase boundaries to construct phase-wise FCs $\{A_t\}_{t=1}^{W}$, with each $A_t\in\mathbb{R}^{N\times N}$. Each $A_t$ is decomposed into $A_t^{+}$/$A_t^{-}$ via a Graph Structure Generator and encoded by a shared Structure-Aware Encoder. 
% Attention weights informative phases to form subject embeddings, while contrastive supervision makes them diagnosis-discriminative relative to original-graph embeddings.

\subsection{Adaptive Phase Partition via Time-Invariant Representation}
% Given a subject's rs-fMRI BOLD time series $X\in\mathbb{R}^{T\times N}$, where $T$ denotes the number of time points and $N$ denotes the number of ROIs,
% As shown in Fig.~\ref{method},
% we first construct an overlapping BOLD segment sequence using a sliding window (size $w$, step $s$), where $s=1$ for all tasks, and $w=200$ for the MDD/BD cohorts while $w=40$ for ASD, yielding
% $K=\left\lfloor\frac{T-w}{s}\right\rfloor+1$ segments $\{X_k\}_{k=1}^{K}$ with $X_k\in\mathbb{R}^{w\times N}$.
% To perform boundary learning, we train a temporal autoencoder to obtain $\mathbf{h}_k=f_{\mathrm{enc}}(X_k)\in\mathbb{R}^{d}$, and factorize each segment into time-invariant component $\mathbf{s}_k$ and time-variant component $\mathbf{u}_k$, i.e., $\mathbf{h}_k=[\mathbf{s}_k,\mathbf{u}_k]$.
% %where $\mathbf{s}_k$ captures state-related (time-invariant) patterns and $\mathbf{u}_k$ captures time-variant fluctuations.
% We implement $f_{\mathrm{enc}}$ as a dilated temporal convolutional network (TCN) with stacked 1D convolutions and residual connections.
% A symmetric decoder reconstructs the input segment as $\hat X_k=f_{\mathrm{dec}}(\mathbf{h}_k)$.
% The training objective is:
As shown in Fig.~\ref{method}, we construct an overlapping segment sequence by sliding windowing (size $w$, step $s{=}1$), with $w{=}200$ for MDD/BD and $w{=}40$ for ASD, yielding
$K=\left\lfloor\frac{T-w}{s}\right\rfloor+1$ segments $\{X_k\}_{k=1}^{K}$, $X_k\in\mathbb{R}^{w\times N}$.
To perform boundary learning, we train a temporal autoencoder to encode each segment as $\mathbf{h}_k=f_{\mathrm{enc}}(X_k)\in\mathbb{R}^{d}$ and disentangle $\mathbf{h}_k=[\mathbf{s}_k,\mathbf{u}_k]$ into time-invariant $\mathbf{s}_k$ and time-variant $\mathbf{u}_k$ components.
We implement $f_{\mathrm{enc}}$ as a dilated TCN and use a symmetric decoder to reconstruct $\hat X_k=f_{\mathrm{dec}}(\mathbf{h}_k)$.
The training objective is:
\begin{equation}
\mathcal{L}_{\mathrm{APP}}
=
\mathcal{L}_{\mathrm{recon}}
+\lambda_{\mathrm{smooth}}\mathcal{L}_{\mathrm{smooth}}
+\lambda_{\mathrm{orth}}\mathcal{L}_{\mathrm{orth}},
\end{equation}
where $\mathcal{L}_{\mathrm{recon}}$ enforces reconstruction of BOLD segments,
$\mathcal{L}_{\mathrm{smooth}}$ penalizes abrupt changes of $\mathbf{s}_k$ across segments to encourage stable brain-state trajectories,
and $\mathcal{L}_{\mathrm{orth}}$ encourages disentanglement between $\mathbf{s}_k$ and $\mathbf{u}_k$ to reduce state--noise leakage.
% We set $\lambda_{\mathrm{smooth}}=0.1$ and $\lambda_{\mathrm{orth}}=1.0$ for all tasks, tuned on the validation set.
% With the decoupled $\{\mathbf{s}_k\}$,  a changepoint is detected when $d_k>\tau$, where $d_k=\lVert \mathbf{s}_k-\mathbf{s}_{k-1}\rVert_2^2$.
% %Changepoints are detected on $\{\mathbf{s}_k\}$ using $d_k=\lVert \mathbf{s}_k-\mathbf{s}_{k-1}\rVert_2^2$, and a changepoint is declared when $d_k>\tau$.
% The detected changepoints are finally mapped to the original time axis, producing brain-state boundaries
% $0=c_0<c_1<\cdots<c_W=T$ adaptively.
% Changepoints are detected on $\{\mathbf{s}_k\}$ when $d_k>\tau$, where $d_k=\lVert \mathbf{s}_k-\mathbf{s}_{k-1}\rVert_2^2$, yielding boundaries
% $0=c_0<\cdots<c_W=T$.
% Changepoints are detected on $\{\mathbf{s}_k\}$ when $d_k>\tau$ (with $d_k=\lVert \mathbf{s}_k-\mathbf{s}_{k-1}\rVert_2^2$), yielding boundaries $0=c_0<\cdots<c_W=T$.
% Changepoints are detected on $\{\mathbf{s}_k\}$ when $d_k>\tau$ (with $d_k=\lVert \mathbf{s}_k-\mathbf{s}_{k-1}\rVert_2^2$), yielding boundaries on the original BOLD time axis $0=c_0<\cdots<c_W=T$.
% Changepoints are detected on $\{\mathbf{s}_k\}$ when $d_k>\tau$ (with $d_k=\lVert \mathbf{s}_k-\mathbf{s}_{k-1}\rVert_2^2$), yielding $W$ phases with boundaries on the original BOLD time axis $0=c_0<\cdots<c_W=T$,
% where $\tau$ is a changepoint threshold and $\{c_t\}_{t=0}^{W}$ are the detected boundary indices.
Changepoints are detected on $\{\mathbf{s}_k\}$ when $d_k>\tau_{\mathrm{c}}$ (with $d_k=\lVert \mathbf{s}_k-\mathbf{s}_{k-1}\rVert_2^2$), yielding $W$ phases with boundaries on the original BOLD time axis $0=c_0<\cdots<c_W=T$,
where $\tau_{\mathrm{c}}$ is a  threshold and $\{c_t\}_{t=0}^{W}$ are the detected boundary indices.
Each adjacent changepoint pair $(c_{t-1},c_t)$ defines one phase,
and we compute the phase-wise FC sequence $\{A_t\}_{t=1}^{W}$ by Pearson correlation within each phase for subsequent modules.
We set $\lambda_{\mathrm{smooth}}=0.1$ and $\lambda_{\mathrm{orth}}=1.0$ for all tasks, and select  $\tau_{\mathrm{c}}$ from $\{0.01,0.05,0.1,0.5\}$ on the validation set (per dataset).

% Changepoints are detected on $\{\mathbf{s}_k\}$ when $d_k>\tau_{\mathrm{cp}}$ (with $d_k=\lVert \mathbf{s}_k-\mathbf{s}_{k-1}\rVert_2^2$), yielding $W$ phases with boundaries on the original BOLD time axis $0=c_0<\cdots<c_W=T$,
% where $\{c_t\}_{t=0}^{W}$ are the detected boundary indices.

\subsection{Incremental  Graph Structure Generator}
\label{sec:mask_generator}
For each $A_t\in\mathbb{R}^{N\times N}$ ($t=1,\ldots,W$), we learn a graph structure $S_t$ to separate disease-relevant from irrelevant connectivities.
\paragraph{Base and incremental structure.}
We learn a trainable base structure $S_0\in\mathbb{R}^{N\times N}$,  initialized to a small positive constant to avoid biasing the model toward any specific connectivity.
For each phase, an MLP $f_{\theta}$ predicts an incremental update $\Delta S_t\in\mathbb{R}^{N\times N}$ determinbed by a compact temporal descriptor $\mathbf{z}_t\in\mathbb{R}^{3}$, defined as $\mathbf{z}_t=\left[\frac{c_{t-1}+c_t}{2T},\;\frac{c_t-c_{t-1}}{T},\;\lVert A_t-A_{t-1}\rVert_F\right]$, where the three terms encodes normalized location, duration, and FC change magnitude. 
Given  $\mathbf{z}_t$, the structure evolves recursively as $\Delta S_t=f_{\theta}(\mathbf{z}_t)$ and $S_t=S_{t-1}+\alpha_{\Delta}\Delta S_t$.
This incremental design captures progressive topological changes across phases with shared parameters and naturally supports variable-length sequences.
%We define We use these raw BOLD timing cues together with inter-phase FC variation to determine how the incremental structure at phase \(t\) retains connectivities. 
We set $\alpha_{\Delta}=0.01$ and choose the initialization constant ($0.05$) on the validation set.

\paragraph{Differentiable structure learning.}
We adopt a straight-through estimator (STE) to separate disease-related from disease-irrelevant connectivities via end-to-end trainable manner:
$\tilde S_t=\mathbbm{1}(S_t>0.5)$ in the forward pass, with gradients backpropagated through $S_t$.
We then obtain $A_t^{+}=\tilde S_t\odot A_t$ and $A_t^{-}=(\mathbf{1}-\tilde S_t)\odot A_t$, where $\odot$ denotes the element-wise (Hadamard) product; $A_t^{+}$ retains selected disease-related connectivities and $A_t^{-}$ captures the complementary topological patterns.

% \paragraph{Differentiable structure learning.}
% To separate disease-related from disease-irrelevant connectivities, we adopt a straight-through estimator (STE).
% In the forward pass, $\tilde S_t=\mathbbm{1}(S_t>\tau)$, with gradients backpropagated through $S_t$.
% We obtain $A_t^{+}=\tilde S_t\odot A_t$, and define $A_t^{-}=A_t-A_t^{+}$.
% Here $\odot$ denotes the element-wise product; $A_t^{+}$ retains selected disease-related connectivities and $A_t^{-}$ captures the complementary topological information.

% \paragraph{Differentiable structure learning.}
% To explicitly separate disease-related connectivities from disease-irrelevant connectivities by structure learning via end-to-end trainable manner, we adopt a straight-through estimator (STE) method: $\tilde S_t=\mathbbm{1}(S_t>\tau)$ in the forward pass, with gradients backpropagated through $S_t$.
% We then obtain $A_t^{+}=\tilde S_t\odot A_t$ and $A_t^{-}=(\mathbf{1}-\tilde S_t)\odot A_t$, where $\odot$ denotes the element-wise (Hadamard) product, $A_t^{+}$ retains disease-related connectivities selected by the learned structure, and $A_t^{-}$ collects the complementary disease-irrelevant structure.
\paragraph{Structure regularization.}
We regularize the structures with three terms:
a binarization loss $\mathcal{L}_{\mathrm{bin}}=\sum_{t=1}^{W}\lVert S_t\odot(1-S_t)\rVert_F^2$ that encourages near-binary structures,% for clear connectivity-level separation,
a temporal consistency loss $\mathcal{L}_{\mathrm{ms}}=\sum_{t=2}^{W}\log\!\big(1+\exp(\lVert S_t-S_{t-1}\rVert_F^2-\delta)\big)$ that promotes smooth evolution across phases, where the margin $\delta$ allows mild variations but penalizes abrupt changes, and a sparsity loss \(\mathcal{L}_{\mathrm{sp}}=\sum_{t=1}^{W}\frac{1}{N^2}\lVert S_t\rVert_{1}\),  which encourages sparse disease-related topology and is consistent with evidence that large number of connectivities are not statistically associated with disease~\cite{zhang2023brainusl}.
We set $\delta=0.1$ on the validation set.

% \subsection{Structure-Aware Encoder}
% Given $A_t^{(\cdot)}\in\{A_t^{+},A_t^{-},A_t\}$, we extract a phase embedding with a shared Structure-Aware Encoder $g(\cdot)$.
% Specifically, we first apply an edge-to-edge operator
% $E_t^{(\cdot)}=\mathrm{Conv}_{N,1}(A_t^{(\cdot)})+\mathrm{Conv}_{1,N}(A_t^{(\cdot)})$
% to capture cross-row and cross-column dependencies among ROI pairs, which preserves the structured interactions in FC matrices.
% We then aggregate edge features to node-level patterns using an edge-to-node operator
% $F_t^{(\cdot)}=\mathrm{Conv}_{1,N}(E_t^{(\cdot)})$,
% yielding compact node-centric embeddings.
% Finally, we obtain a topology-aware graph embedding by
% $\mathbf{h}_t^{(\cdot)}=g(A_t^{(\cdot)})=\mathrm{MLP}(\mathrm{Flatten}(F_t^{(\cdot)}))\in\mathbb{R}^{d}$, where $d$ denotes the embedding dimension.
% The encoder is shared across $\{A_t^{+},A_t^{-},A_t\}$ to enforce consistent representations and reduce parameters, producing
% $\mathbf{h}_t^{+},\mathbf{h}_t^{-}$, and $\mathbf{h}_t^{0}$, respectively.

\subsection{Structure-Aware Encoder}
For each $A_t^{(\cdot)}\in\{A_t^{+},A_t^{-},A_t\}$, we compute a phase embedding using a shared Structure-Aware Encoder $g(\cdot)$.
We first apply an edge-to-edge operator
$E_t^{(\cdot)}=\mathrm{Conv}_{N,1}(A_t^{(\cdot)})+\mathrm{Conv}_{1,N}(A_t^{(\cdot)})$
to capture structured interactions among ROI pairs, and then use an edge-to-node operator
$F_t^{(\cdot)}=\mathrm{Conv}_{1,N}(E_t^{(\cdot)})$ to aggregate edge features.
Finally, we obtain a topology-aware embedding
$\mathbf{h}_t^{(\cdot)}=g(A_t^{(\cdot)})=\mathrm{MLP}(\mathrm{Flatten}(F_t^{(\cdot)}))\in\mathbb{R}^{d}$, where $d$ denotes the embedding dimension.
The encoder is shared across $\{A_t^{+},A_t^{-},A_t\}$ to enforce consistent representations and reduce parameters, producing
$\mathbf{h}_t^{+},\mathbf{h}_t^{-}$, and $\mathbf{h}_t^{0}$, respectively.

\subsection{Spatio-Temporal Contrastive Learning}

\paragraph{Attention and aggregation.}
Given phase embeddings $\{\mathbf{h}_t^{+},\mathbf{h}_t^{-},\mathbf{h}_t^{0}\}_{t=1}^{W}$ from the shared subgraph encoder, we assign each phase an importance weight with
$\alpha_t^{(\cdot)}=
\frac{\exp\!\big(f_{\mathrm{attn}}(\mathbf{h}_t^{(\cdot)})\big)}
{\sum_{k=1}^{W}\exp\!\big(f_{\mathrm{attn}}(\mathbf{h}_k^{(\cdot)})\big)}$, $(\cdot)\in\{+,-,0\}$,
where $f_{\mathrm{attn}}$ is implemented as a two-layer MLP with a Tanh nonlinearity (Linear--Tanh--Linear) that maps an embedding to a scalar score.
We identify important phases using $\alpha_t^{+}$ by $\mathcal{I}=\{t\mid \alpha_t^{+}>1/W\}$, or $\mathcal{I}=\{\arg\max_t \alpha_t^{+}\}$ if empty, and normalize $\tilde{\alpha}_t^{+}=\alpha_t^{+}\big/\sum_{k\in\mathcal{I}}\alpha_k^{+}$.
We then aggregate disease-related connectivity from important phases as $\mathbf{H}^{++}=\sum_{t\in\mathcal{I}}\tilde{\alpha}_t^{+}\mathbf{h}_t^{+}$.
Meanwhile, the original graph embedding $\mathbf{H}^{0}=\sum_{t=1}^{W}\alpha_t^{0}\mathbf{h}_t^{0}$ is obtained for contrastive supervision. For the complementary representation, we  define $\mathbf{H}^{-}=\sum_{t=1}^{W}\alpha_t^{-}\mathbf{h}_t^{-}$, which is later aligned with $\mathbf{H}^{0}$ via a reference-guided contrastive term.
The classifier predicts $\hat{\mathbf{y}}=\mathrm{softmax}(W_c\mathbf{H}^{++}+\mathbf{b}_c)$,
where $W_c$ and $\mathbf{b}_c$ are the learnable weight matrix and bias of the final linear classifier.

\paragraph{Spatio-temporal contrastive loss.}
With the original graph embedding $\mathbf{H}^{0}$ as reference, we define an InfoNCE-style objective with two terms, $\mathcal{L}_{\mathrm{ref}}$ and $\mathcal{L}_{\mathrm{usl}}$:
\begin{equation}
\mathcal{L}_{\mathrm{str}}
=
w_{\mathrm{ref}}\mathcal{L}_{\mathrm{ref}}
+
w_{\mathrm{usl}}\mathcal{L}_{\mathrm{usl}} ,
\label{eq:contra}
\end{equation}

\begin{equation}
\mathcal{L}_{\mathrm{ref}}
=
-\sum_{i}\frac{1}{|\mathcal{P}(i)|}\sum_{p\in\mathcal{P}(i)}
\log\frac{\exp(\ell^{\mathrm{ref}}_{ip})}{\sum_{j\neq i}\exp(\ell^{\mathrm{ref}}_{ij})},
\label{eq:st_infonce_ref}
\end{equation}

\begin{equation}
\mathcal{L}_{\mathrm{usl}}
=
-\sum_{i}\log\frac{\exp(\ell^{\mathrm{usl}}_{ii})}{\sum_{j}\exp(\ell^{\mathrm{usl}}_{ij})}.
\label{eq:st_infonce_usl}
\end{equation}
where $\mathcal{L}_{\mathrm{ref}}$ measures the semantic similarity gain \textit{wrt}. the disease-related spatio-temporal patterns by removing the inherent similarity estimated by $\mathbf{H}^{0}$, and $\mathcal{L}_{\mathrm{usl}}$ aligns the disease-irrelevant spatio-temporal representation $\mathbf{H}^{-}$ with $\mathbf{H}^{0}$, so $\mathbf{H}^{-}$ remains stable and forces $\mathbf{H}^{++}$ to faster convergence during training. 
This is critical because $\mathbf{H}^{++}$ and $\mathbf{H}^{-}$ are jointly learned from complementary structures ($A_t^{+}$ and $A_t^{-}$).
% \begin{equation}
% \mathcal{L}_{\mathrm{ref}}
% =
% -\sum_{i}\frac{1}{|\mathcal{P}(i)|}\sum_{p\in\mathcal{P}(i)}
% \log\frac{\exp(\ell^{\mathrm{ref}}_{ip})}{\sum_{j\neq i}\exp(\ell^{\mathrm{ref}}_{ij})},
% \label{eq:st_infonce_ref}
% \end{equation}
% \begin{equation}
% \mathcal{L}_{\mathrm{usl}}
% =
% -\sum_{i}\log\frac{\exp(\ell^{\mathrm{usl}}_{ii})}{\sum_{j}\exp(\ell^{\mathrm{usl}}_{ij})}.
% \label{eq:st_infonce_usl}
% \end{equation}
% Together, $\mathcal{L}_{\mathrm{ref}}$ and $\mathcal{L}_{\mathrm{usl}}$ encourage $\mathbf{H}^{++}$ to focus on diagnosis-relevant spatio-temporal patterns. We use temperature-scaled similarity scores:
% \begin{align}
% \ell^{\mathrm{ref}}_{ij}
% &=
% \frac{\mathrm{sim}(\mathbf{H}^{++}_i,\mathbf{H}^{++}_j)
% -\beta\,\mathrm{sim}(\mathbf{H}^{0}_i,\mathbf{H}^{0}_j)}{\tau}, \label{eq:ref_logit_compact}\\
% \ell^{\mathrm{usl}}_{ij}
% &=
% \frac{\mathrm{sim}(\mathbf{H}^{0}_i,\mathbf{H}^{-}_j)}{\tau}, \label{eq:usl_logit_compact}
% \end{align}
% where $\mathcal{P}(i)=\{p\neq i \mid y_p=y_i\}$ denotes same-label positives of subject $i$ within a minibatch, $\mathrm{sim}(\cdot,\cdot)$ denotes cosine similarity, $\tau$ is the temperature,  and $\beta$ controls the reference strength.
Together, $\mathcal{L}_{\mathrm{ref}}$ and $\mathcal{L}_{\mathrm{usl}}$ encourage $\mathbf{H}^{++}$ to focus on diagnosis-relevant spatio-temporal patterns.
We use temperature-scaled similarity scores, with
$\ell^{\mathrm{ref}}_{ij}=\big(\mathrm{sim}(\mathbf{H}^{++}_i,\mathbf{H}^{++}_j)-\beta\,\mathrm{sim}(\mathbf{H}^{0}_i,\mathbf{H}^{0}_j)\big)/\tau$.
Similarly,
$\ell^{\mathrm{usl}}_{ij}=\mathrm{sim}(\mathbf{H}^{0}_i,\mathbf{H}^{-}_j)/\tau$.
Here $\mathcal{P}(i)=\{p\neq i \mid y_p=y_i\}$ denotes same-label positives of subject $i$ within a minibatch, $\mathrm{sim}(\cdot,\cdot)$ denotes cosine similarity, $\tau$ is the temperature,  and $\beta$ controls the reference strength.
We set $w_{\mathrm{ref}}=w_{\mathrm{usl}}=1.0$, $\tau=0.1$, and $\beta=0.65$ on the validation set.

\subsection{Loss and Training}
\label{sec:loss}
We first pretrain APP with $\mathcal{L}_{\mathrm{APP}}$ to infer state-consistent phase boundaries via changepoint detection, and then optimize the remaining modules end-to-end with the following objective:

\begin{equation}
\mathcal{L}
=
\min\;
\underbrace{\mathcal{L}_{\mathrm{ce}}+\lambda_{\mathrm{str}}\mathcal{L}_{\mathrm{str}}}_
{\makebox[0pt][c]{\scriptsize\shortstack{Diagnosis-label\\supervision}}}
\mathbin{+}
\underbrace{\lambda_{\mathrm{bin}}\mathcal{L}_{\mathrm{bin}}
+\lambda_{\mathrm{ms}}\mathcal{L}_{\mathrm{ms}}
+\lambda_{\mathrm{sp}}\mathcal{L}_{\mathrm{sp}}}_
{\makebox[0pt][c]{\scriptsize Structure regularization}},
\label{eq:total_loss}
\end{equation}
% \begin{equation}
% \mathcal{L}
% =
% \underbrace{\mathcal{L}_{\mathrm{ce}}+\lambda_{\mathrm{str}}\mathcal{L}_{\mathrm{str}}}_{\parbox{0.5cm}{\centering \text{Diagnosis-label}\\\text{supervision}}}
% +
% \underbrace{\lambda_{\mathrm{bin}}\mathcal{L}_{\mathrm{bin}}
% +\lambda_{\mathrm{ms}}\mathcal{L}_{\mathrm{ms}}
% +\lambda_{\mathrm{sp}}\mathcal{L}_{\mathrm{sp}}}_{\text{Structure regularization}},
% \label{eq:total_loss}
% \end{equation}
%We group the remaining terms into a \emph{diagnosis-label supervision} $\mathcal{L}_{\mathrm{ce}}+\lambda_{\mathrm{str}}\mathcal{L}_{\mathrm{str}}$, which directly optimizes diagnosis and enhances label-consistent discrimination, and a \emph{structure regularization} term $\lambda_{\mathrm{bin}}\mathcal{L}_{\mathrm{bin}}+\lambda_{\mathrm{ms}}\mathcal{L}_{\mathrm{ms}}+\lambda_{\mathrm{sp}}\mathcal{L}_{\mathrm{sp}}$, which encourages stable and interpretable disease-related topology. %We set $\lambda_{\mathrm{str}}=1.0$, $\lambda_{\mathrm{bin}}=0.5$, $\lambda_{\mathrm{ms}}=0.8$, and $\lambda_{\mathrm{sp}}=0.05$ 
% where $\mathcal{L}_{\mathrm{ce}}$ is the cross-entropy loss, $\lambda_{\mathrm{str}}$, $\lambda_{\mathrm{bin}}$, $\lambda_{\mathrm{ms}}$, and $\lambda_{\mathrm{sp}}$  are determined  via a lightweight grid search on the validation set to balance the contributions of all loss terms.
where $\mathcal{L}_{\mathrm{ce}}$ is the cross-entropy loss.
We set $\lambda_{\mathrm{str}}=1.0$, $\lambda_{\mathrm{bin}}=1.0$, $\lambda_{\mathrm{ms}}=1.0$, and $\lambda_{\mathrm{sp}}=0.27$, tuned on the validation set.

% We optimize the remaining modules with a task objective and structure regularization, and set $\lambda_{\mathrm{str}}=1.0$, $\lambda_{\mathrm{bin}}=0.5$, $\lambda_{\mathrm{ms}}=0.8$, and $\lambda_{\mathrm{sp}}=0.05$ via a lightweight grid search.

\section{Experiments and Results}
\subsection{Dataset and Experimental Details}
We evaluate BrainSTR on two rs-fMRI benchmarks for three diagnoses (\textbf{MDD}, \textbf{BD}, \textbf{ASD}). The private cohort\footnote{Approved by the Medical Science Ethics Committee of **** University (Ref. ****).} includes \textbf{246} controls, \textbf{151} MDD, and \textbf{126} BD subjects, preprocessed via \textbf{DPABI}~\cite{yan2016dpabi}. From the public \textbf{ABIDE} dataset, we use only the \textbf{NYU} site (\textbf{74} ASD, \textbf{98} controls) to reduce multi-center variability.
We used AAL-based pre-processed functional data for evaluation. 
Five-fold  cross-validation is applied, with hyperparameters tuned on internal validation set. 

% We evaluate BrainSTR on two rs-fMRI benchmarks for three diagnosis tasks (\textbf{MDD}, \textbf{BD}, \textbf{ASD}). 
% The private cohort\footnote{Approved by the Medical Science Ethics Committee of **** University (Ref. ****).} contains \textbf{246} controls, \textbf{151} MDD, and \textbf{126} BD subjects, preprocessed with \textbf{DPABI}~\cite{yan2016dpabi}. 
% For \textbf{ABIDE}, we use only the \textbf{NYU} site (\textbf{74} ASD, \textbf{98} controls) to reduce multi-center variability. 
% All experiments use AAL-based preprocessed fMRI data. We perform five-fold cross-validation and tune hyperparameters on internal validation splits.

\setlength{\tabcolsep}{4pt}
\begin{table*}[!h]
\centering
\caption{Comparison with state-of-the-art methods on MDD, BD, and ASD diagnosis. The best results are in bold. The average and std of ACC and AUC are reported.}

\label{tab:cd_gd}
\resizebox{\textwidth}{!}{
\begin{tabular}{llcccccccc}
\toprule
\multirow{2}{*}{Category} & \multirow{2}{*}{Method} 
& \multicolumn{2}{c}{HC vs. MDD} 
& \multicolumn{2}{c}{HC vs. BD} 
& \multicolumn{2}{c}{HC vs. ASD (NYU center)} \\
\cmidrule(lr){3-4} \cmidrule(lr){5-6} \cmidrule(lr){7-8}
& & ACC(\%) & AUC(\%) & ACC(\%) & AUC(\%) & ACC(\%) & AUC(\%) \\
\midrule
\multirow{3}{*}{Traditional}
& FC + SVM
& $68.4\pm2.7$ & $67.2\pm3.2$
& $65.6\pm3.1$ & $67.3\pm3.5$
& $68.1\pm2.9$ & $69.7\pm3.6$ \\
& FC + RF
& $64.7\pm2.9$ & $61.2\pm3.4$
& $65.2\pm3.1$ & $63.6\pm4.0$
& $66.2\pm2.8$ & $65.1\pm3.7$ \\
& SVM-RFE \cite{guyon2002gene}
& $70.3\pm3.8$ & $70.1\pm4.0$
& $66.9\pm3.7$ & $67.6\pm3.9$
& $65.2\pm4.1$ & $65.9\pm4.3$ \\
\midrule

\multirow{7}{*}{Static graph}
& GroupINN \cite{yan2019groupinn}
& $67.1\pm3.9$ & $65.7\pm4.2$
& $68.3\pm3.8$ & $65.2\pm4.0$
& $63.3\pm4.3$ & $64.3\pm4.4$ \\
& BrainGNN \cite{li2021braingnn}
& $65.2\pm3.6$ & $63.7\pm3.9$
& $69.1\pm3.4$ & $68.7\pm3.6$
& $65.2\pm3.8$ & $65.4\pm4.0$ \\
& MVS-GCN \cite{wen2022mvs}
& $68.1\pm3.5$ & $68.1\pm3.7$
& $67.1\pm3.6$ & $65.0\pm3.8$
& $69.3\pm3.7$ & $68.3\pm3.9$ \\
& ASD-DiagNet \cite{eslami2019asd}
& $68.2\pm3.4$ & $68.0\pm3.6$
& $72.1\pm3.2$ & $72.1\pm3.4$
& $67.4\pm3.5$ & $67.5\pm3.7$ \\
& FRL \cite{kang2023learnable}
& $68.0\pm3.3$ & $68.9\pm3.5$
& $71.1\pm3.2$ & $70.7\pm4.1$
& $67.1\pm3.5$ & $69.4\pm3.7$ \\
& EAG-RS \cite{jung2023eag}
& $66.5\pm3.6$ & $70.9\pm3.8$
& $73.5\pm3.1$ & $72.1\pm4.3$
& $66.8\pm3.7$ & $68.1\pm3.9$ \\
& D-CoRP \cite{hu2024d}
& $71.7\pm2.7$ & $71.0\pm2.9$
& $72.6\pm2.7$ & $71.8\pm2.9$
& $67.9\pm1.7$ & $65.3\pm2.1$ \\
\midrule

\multirow{5}{*}{Dynamic graph}
& MDGL \cite{ma2023multi}
& $67.4\pm3.7$ & $66.7\pm3.9$
& $69.1\pm3.5$ & $70.3\pm3.7$
& $66.9\pm3.9$ & $64.1\pm4.1$ \\
& STA-GIN \cite{kim2021learning}
& $73.1\pm3.2$ & $66.0\pm3.5$
& $72.4\pm3.1$ & $68.5\pm3.3$
& $68.9\pm3.4$ & $67.2\pm3.7$ \\
& BrainDGT \cite{shehzad2025dynamic}
& $74.3\pm2.6$ & $71.3\pm2.8$
& $73.7\pm3.4$ & $74.7\pm2.7$
& $69.1\pm3.1$ & $69.7\pm4.2$ \\
& MCDGLN \cite{wang2025mcdgln}
& $72.6\pm2.5$ & $71.1\pm3.7$
& $71.6\pm3.2$ & $72.1\pm2.6$
& $67.2\pm2.7$ & $69.3\pm2.9$ \\
& ST-GCN \cite{gadgil2020spatio}
& $60.3\pm4.8$ & $62.9\pm5.1$
& $61.3\pm4.7$ & $60.9\pm4.9$
& $56.7\pm5.2$ & $52.6\pm5.6$ \\
\midrule

\multirow{6}{*}{Ours}
& w/o APP
& $74.3\pm3.3$ & $77.2\pm3.5$
& $75.9\pm3.2$ & $76.2\pm3.4$
& $70.1\pm3.5$ & $70.4\pm3.7$ \\
& w/o $\Delta S$
& $68.6\pm4.4$ & $64.3\pm4.7$
& $69.4\pm4.3$ & $66.8\pm4.6$
& $69.4\pm4.6$ & $66.8\pm4.9$ \\
& w/o $L_{\text{str}}$
& $72.4\pm3.8$ & $77.3\pm4.0$
& $72.4\pm3.7$ & $74.5\pm3.9$
& $71.3\pm4.0$ & $72.7\pm4.2$ \\
& w/o $L_{\text{ms}}$
& $74.3\pm3.6$ & $71.3\pm3.8$
& $73.7\pm3.5$ & $70.2\pm3.7$
& $71.7\pm3.8$ & $70.2\pm4.0$ \\
& w/o $L_{\text{bin}}$
& $76.7\pm3.1$ & $75.2\pm3.3$
& $77.6\pm3.0$ & $78.2\pm3.2$
& $72.1\pm3.3$ & $71.8\pm3.5$ \\
& w/o $L_{\text{sp}}$
& $76.2\pm3.0$ & $76.7\pm3.2$
& $77.8\pm2.9$ & $79.0\pm3.1$
& $72.1\pm3.2$ & $72.5\pm3.4$ \\
\midrule
\textbf{Full Model} & \textbf{BrainSTR}
& \textbf{77.2$\pm$1.6} & \textbf{77.8$\pm$2.8}
& \textbf{78.2$\pm$1.5} & \textbf{79.6$\pm$1.7}
& \textbf{72.4$\pm$1.9} & \textbf{73.0$\pm$2.1} \\
\bottomrule
\end{tabular}%
}
\end{table*}

\subsection{Classification Results}
\label{sec:results}

% \textbf{Comparison with state-of-the-art.}
% As shown in Table~\ref{tab:cd_gd}, BrainSTR achieves the best overall performance on all three diagnoses.
% On the private cohort, BrainSTR reaches \textbf{77.2$\pm$1.6 / 77.8$\pm$2.8} (ACC/AUC) for MDD and
% \textbf{78.2$\pm$1.5 / 79.6$\pm$1.7} for BD, outperforming the strongest dynamic baseline (BrainDGT) by
% \textbf{+2.9/+6.5} and \textbf{+4.5/+4.9} points, respectively.
% On ABIDE (NYU), BrainSTR also yields the best results with \textbf{72.4$\pm$1.9} ACC and \textbf{73.0$\pm$2.1} AUC.
% These gains suggest that learning diagnosis-relevant spatio-temporal evidence instead of aggregating whole-series information provides a more discriminative and robust representation for disorder identification.
% \textbf{Comparison with state-of-the-art} Table~\ref{tab:cd_gd} compares BrainSTR with traditional classifiers built on handcrafted FC features, static-graph models on aggregated FC, and dynamic-graph methods for dFC, where \textbf{BrainDGT} is the strongest dynamic baseline.

\subsubsection{Comparison with state-of-the-art.}
% As summarized in Table~\ref{tab:cd_gd}, we benchmark BrainSTR against representative pipelines from three widely-used paradigms in rs-fMRI diagnosis: \emph{(i) traditional classifiers} built upon handcrafted FC features (e.g., SVM-RFE), \emph{(ii) static-graph learning} that aggregates the whole time series into a single FC graph for graph representation learning, and \emph{(iii) dynamic-graph modeling} that explicitly captures temporal variations of dFC. This coverage enables a fair and comprehensive comparison, where dynamic-graph methods are the most competitive counterparts due to their closest alignment with our problem setting (among which BrainDGT is the strongest baseline in our experiments).
% As summarized in Table~\ref{tab:cd_gd}, we compare BrainSTR with representative approaches from three mainstream paradigms in rs-fMRI diagnosis: \emph{traditional} pipelines using handcrafted FC features, \emph{static-graph} models built on a single aggregated FC graph, and \emph{dynamic-graph} methods that model dFC over time. This design provides a comprehensive and competitive comparison; among dynamic baselines, \textbf{BrainDGT} is the strongest competitor in our experiments.
Table~\ref{tab:cd_gd} compares BrainSTR with traditional classifiers built on handcrafted FC features, static-graph models on aggregated FC, and dynamic-graph models for dFC, where \textbf{BrainDGT} is the strongest dynamic baseline.
On the private cohort, BrainSTR achieves $77.2\pm1.6/77.8\pm2.8$ (ACC/AUC, in \%) for MDD and $78.2\pm1.5/79.6\pm1.7$ for BD, surpassing BrainDGT by $+2.9\%/+6.5\%$ and $+4.5\%/+4.9\%$, respectively. On ABIDE (NYU), BrainSTR remains the top-performing method. Overall, BrainSTR achieves the best performance in terms of both ACC and AUC across all three diseases, indicating robust cross-task superiority and generalization.
We attribute the improvements to  BrainSTR's explicit extraction of diagnosis-relevant spatio-temporal patterns: It learns stable brain-state trajectories via Adaptive Phase Partition (APP), incrementally estimates phase-wise structures to separate disease-related from irrelevant connectivity, and refines discriminative representations through  spatio-temporal contrastive learning.

\textbf{Ablation analysis.}
We further verify the contribution of each key component in Table~\ref{tab:cd_gd}.
Removing \textbf{APP} degrades performance consistently, indicating that adaptive boundary learning provides a better temporal partition  for downstream FC construction and diagnosis.
Discarding the transition cue \textbf{$\Delta S$} causes the largest drop,
highlighting that modeling incremental structure is crucial for capturing diagnosis-relevant dynamic connectivity.
In addition, removing $\mathcal{L}_{\mathrm{str}}$ or individual structure regularizers
also leads to noticeable degradation, 
% demonstrating that the proposed objectives are complementary: 
% they jointly suppress redundant connectivity  patterns while promoting diagnosis-relevant spatio-temporal evidence for classification.
% showing that these objectives are complementary, suppressing redundant connectivity patterns and promoting diagnosis-relevant spatio-temporal evidence.
showing that these objectives are complementary in suppressing redundant connectivity patterns and promoting diagnosis-relevant spatio-temporal evidence.
% \textbf{Ablation analysis.}
% We  verify the contribution of each component.
% Removing \textbf{APP} consistently degrades performance, suggesting that time-invariant boundary learning yields better phase partitions for FC construction.
% Discarding the transition cue \textbf{$\Delta S$} causes the largest drop (e.g., MDD ACC/AUC: 77.2/77.8 $\rightarrow$ 68.6/64.3), showing the importance of incremental structure modeling.
% Removing $\mathcal{L}_{\mathrm{str}}$ or individual regularizers ($\mathcal{L}_{\mathrm{bin}}, \mathcal{L}_{\mathrm{ms}}, \mathcal{L}_{\mathrm{sp}}$) also hurts performance, indicating complementary objectives that suppress redundancy and enhance diagnosis-relevant spatio-temporal evidence.

\begin{figure}[t]
\centering
\begin{minipage}[t]{0.49\textwidth}
  \centering
  \includegraphics[width=\linewidth]{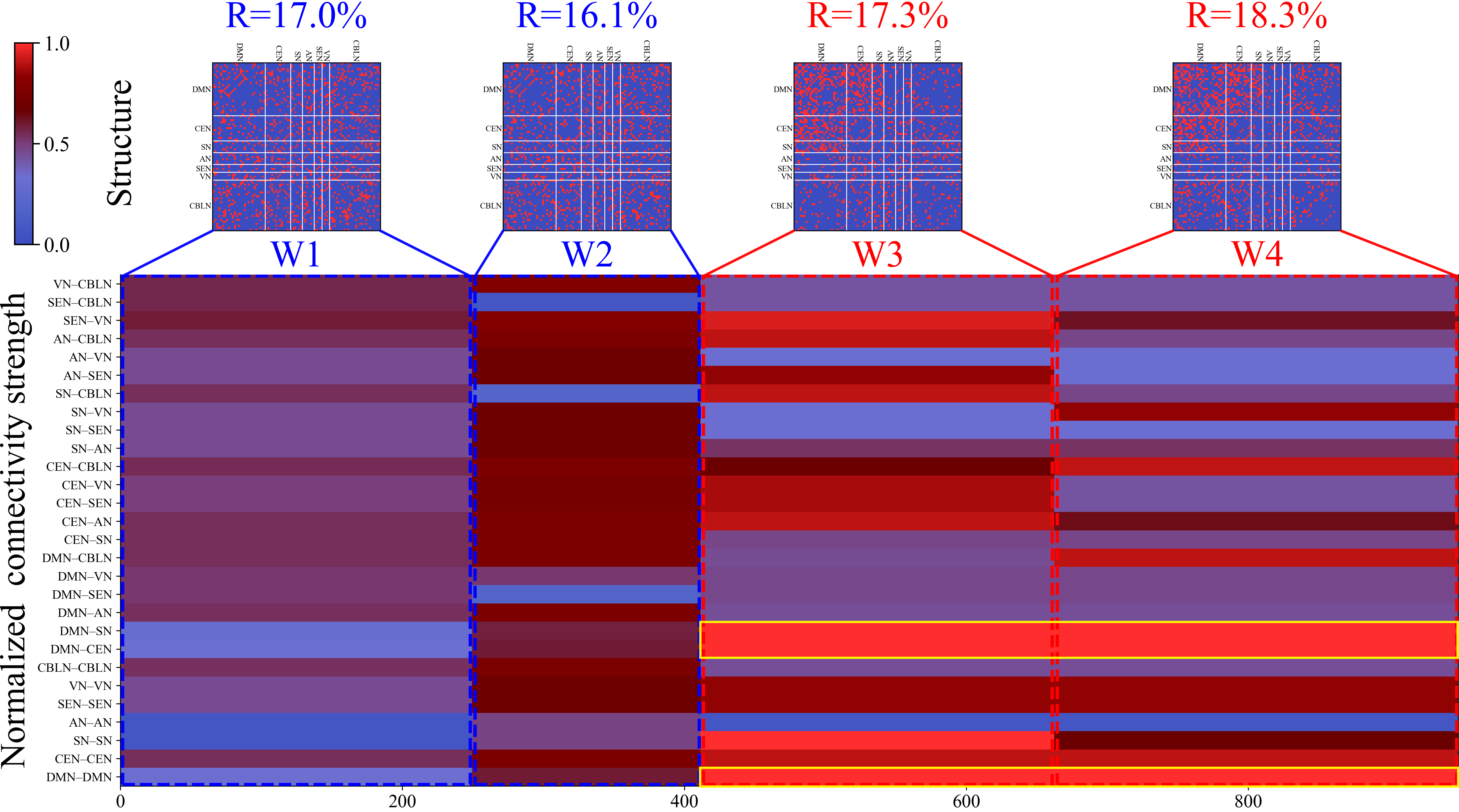}
  \small (a) Representative subject \#1
\end{minipage}\hfill
\begin{minipage}[t]{0.49\textwidth}
  \centering
  \includegraphics[width=\linewidth]{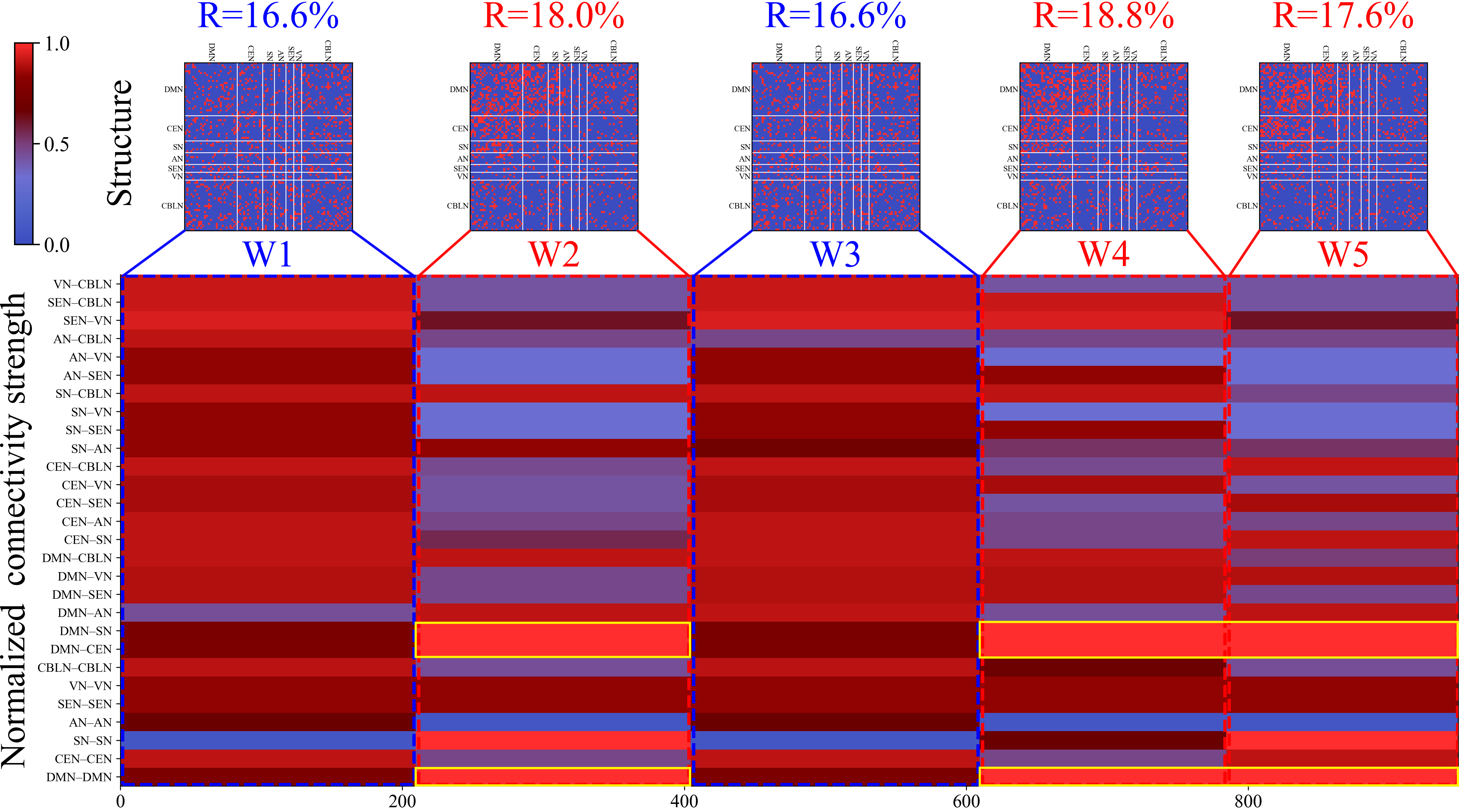}
  \small (b) Representative subject \#2
\end{minipage}

% \caption{Case-level temporal--topological interpretability of BrainSTR on representative subjects. Top: retained-connectivity ratio $R$  in each phase. Bottom: normalized  connectivity strength of subnetworks, with important phases in red and non-important phases in blue; stable co-activation is highlighted in yellow.}
% \caption{Case-level temporal--topological interpretability of BrainSTR on representative subjects. Top: phase-wise structures, together with the overall retained-connectivity ratio $R$ of each phase. Bottom: normalized  connectivity strength of subnetworks, with important phases in red and non-important phases in blue; consistent DMN-centered patterns are highlighted in yellow.}

\caption{Case-level temporal--topological interpretability of BrainSTR on representative subjects. Top: phase-wise structures, together with the overall retained-connectivity ratio $R$ of each phase. Bottom: normalized  connectivity strength of subnetworks.}

\label{inter}
\end{figure}

\begin{figure*}[!t]
    \centering
    \setlength{\tabcolsep}{1.2pt}
    \renewcommand{\arraystretch}{1.0}

    % 可调参数：左侧行标签整体上移量（想再上移就调大）
    \newcommand{\RowLabelRaise}{4ex}

    \begin{tabular}{@{}c c c c@{}}
        & \textbf{DMN--DMN} & \textbf{DMN--CEN} & \textbf{DMN--SN} \\[0.5mm]

        \raisebox{\RowLabelRaise}[0pt][0pt]{\rotatebox[origin=c]{90}{\textbf{Imp}}} &
        \includegraphics[width=0.305\textwidth]{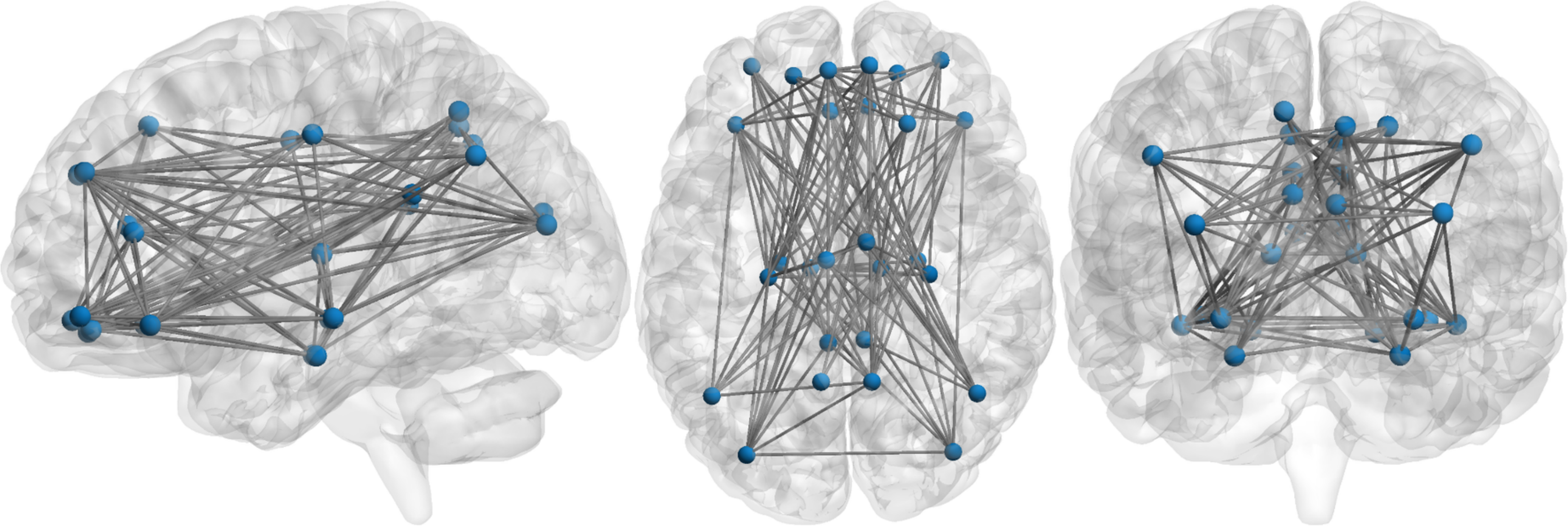} &
        \includegraphics[width=0.305\textwidth]{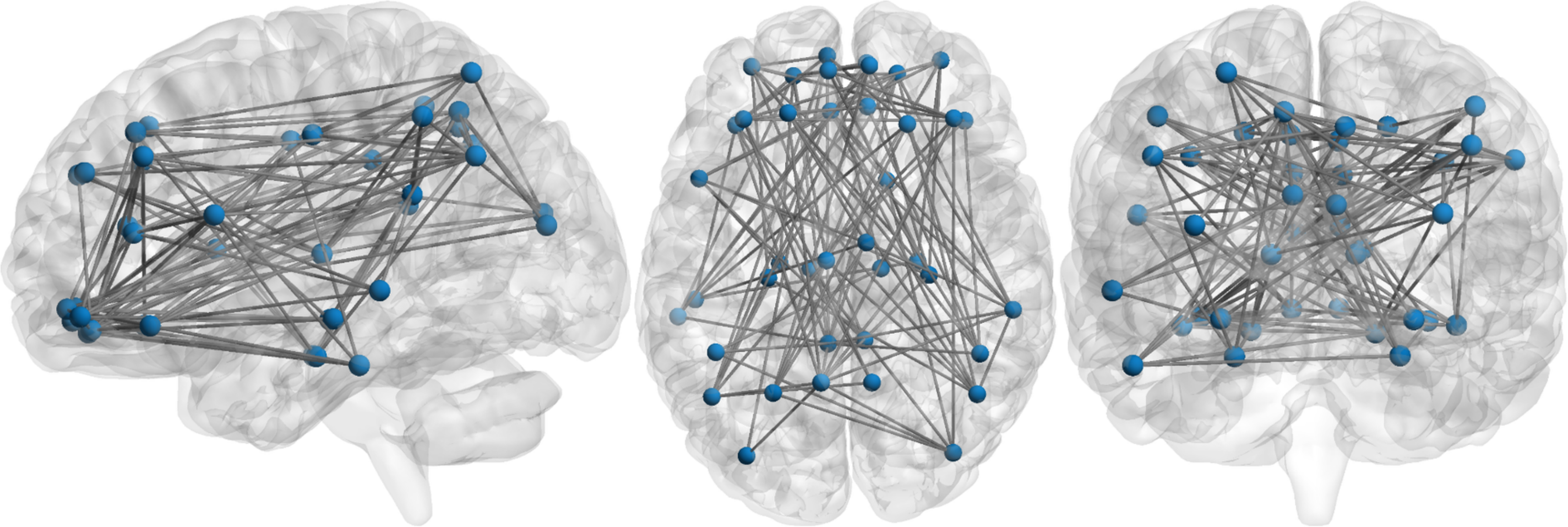} &
        \includegraphics[width=0.305\textwidth]{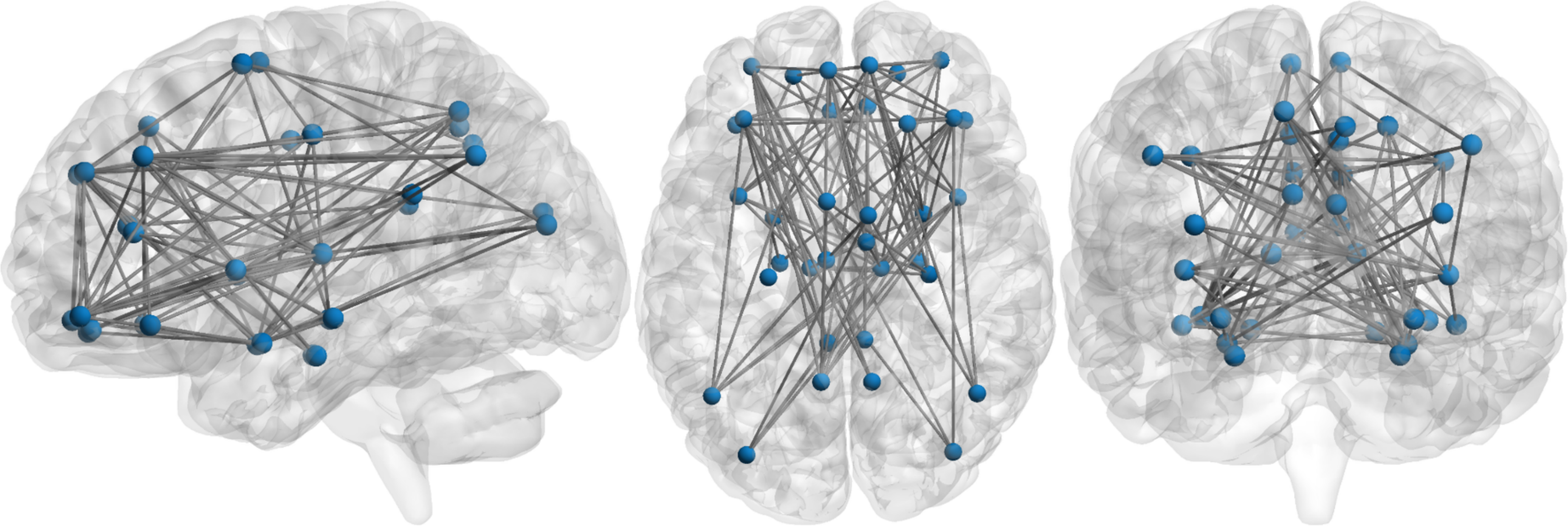} \\[-0.8mm]

        \raisebox{\RowLabelRaise}[0pt][0pt]{\rotatebox[origin=c]{90}{\textbf{Non}}} &
        \includegraphics[width=0.305\textwidth]{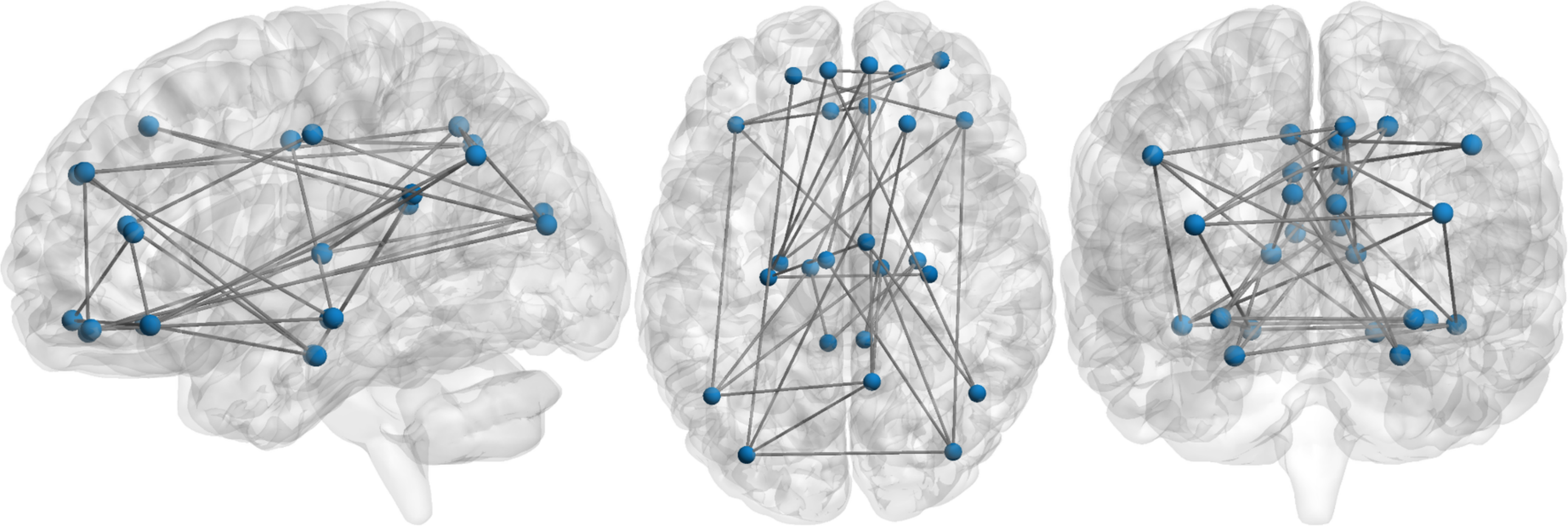} &
        \includegraphics[width=0.305\textwidth]{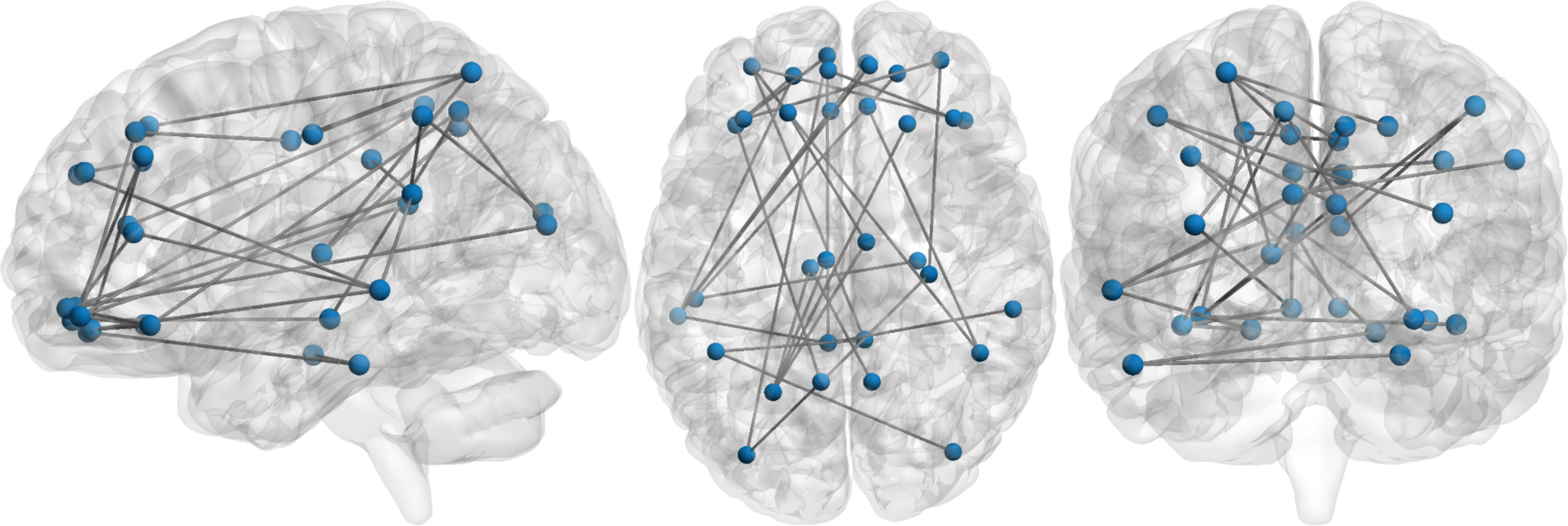} &
        \includegraphics[width=0.305\textwidth]{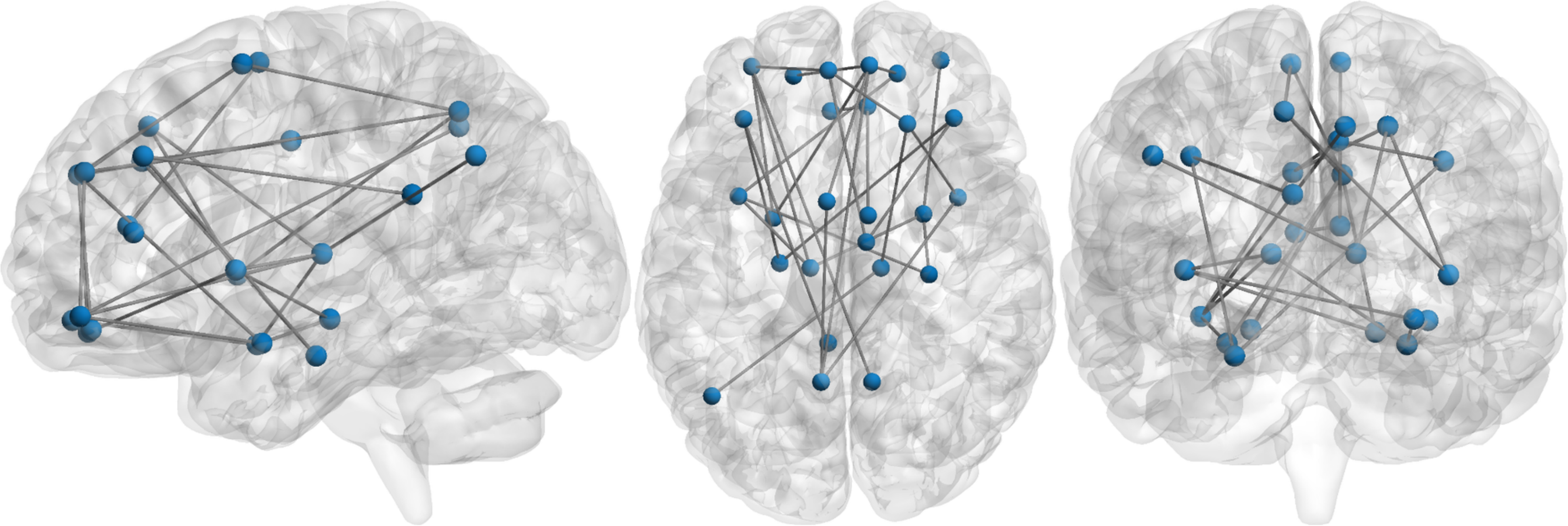} \\
    \end{tabular}

    \vspace{-1mm}
    % \caption{\textbf{Group-level DMN-centered structure retention.}
    % Cohort-aggregated mask-retained connectivity restricted to three DMN-centered subnetwork pairs.
    % Top: important phases; bottom: non-important phases.}
        % \caption{Group-level DMN-centered structure retention. Columns show retained structures for DMN--DMN, DMN--CEN, and DMN--SN pairs. For each pair, the top row (Imp) and bottom row (Non) visualize the group-aggregated retained connectivities in important and non-important phases, respectively, shown from three brain views.}
        \caption{Group-level DMN-centered structure retention. The top row (Imp) and bottom row (Non) visualize the group-aggregated retained connectivities in important and non-important phases, respectively, shown from three brain views.}

    \label{fig:group_subnet_brainnet}
\end{figure*}

\subsection{Interpretability}
\textbf{Individual-level.} On the MDD cohort, important phases (red) show higher $R$ with denser DMN--DMN/DMN--CEN/DMN--SN retention than non-important phases (blue) (Fig.~\ref{inter}). 
The normalized  connectivity strength over subnetwork pairs consistently highlights these DMN-related pairs across important phases (yellow), matching prior findings on altered DMN interactions~\cite{manoliu2014insular,yan2019reduced,tozzi2021reduced,kaiser2015large}.

% \textbf{Individual-level.} Using MDD as an example, representative subjects show stable DMN-centered evidence (Fig.~\ref{inter}). 
% \emph{Top:} important phases (red) preserve denser DMN--DMN/DMN--CEN/DMN--SN connectivity, yielding higher retained-connectivity ratio $R$ than non-important phases (blue). 
% \emph{Bottom:}  the normalized (phase-wise, $[0,1]$) connectivity strength of subnetworks (visualized as a heatmap) consistently emphasizes these DMN-related subnetwork pairs across important phases (yellow), matching prior findings on altered DMN interactions~\cite{manoliu2014insular,yan2019reduced,tozzi2021reduced,kaiser2015large}.

% \textbf{Population-level.} The retained structures are denser in important phases, while non-important phases remain sparse and diffuse (Fig.~\ref{fig:group_subnet_brainnet}); 

\textbf{Population-level.} On the MDD cohort, the retained structures are denser in important phases, while non-important phases remain sparse and diffuse (Fig.~\ref{fig:group_subnet_brainnet}).

% BD and ASD show similar trends, supporting that BrainSTR captures diagnosis-relevant evidence both temporally and topologically.
The BD cohort and ABIDE (NYU) show similar trends, supporting that BrainSTR captures diagnosis-relevant evidence both temporally and topologically.

\section{Conclusion}
% \textcolor{red}{In this work, we addressed a key limitation of dynamic brain-network diagnosis: disease-related evidence is sparse across both time and topology, while irrelevant fluctuations are pervasive.  
% We proposed \textbf{BrainSTR}, which jointly performs adaptive phase partition, attention-based critical-phase selection, and structure-based topological filtering, and optimizes these components with reference-adjusted contrastive supervision.  
% This design shifts representation learning from whole-sample aggregation to \emph{disease-relevant spatio-temporal evidence}, yielding a more structured and interpretable embedding space.  
% Experiments on ASD, BD, and MDD show consistent performance gains over strong baselines.  
% Importantly, the discovered critical phases and subnetworks are neurobiologically plausible and align with prior findings, supporting the clinical interpretability of BrainSTR.}
% In this work, we presented BrainSTR for interpretable dynamic brain-network diagnosis. By combining Adaptive Phase Partition, critical-phase selection, structure-based topological filtering, and reference-adjusted contrastive supervision, BrainSTR shifts learning from whole-sample aggregation to diagnosis-relevant spatio-temporal evidence. Experiments on ASD, BD, and MDD show consistent improvements over strong baselines. The identified critical phases and subnetworks are neurobiologically plausible, supporting the clinical interpretability of the proposed framework.
In this work, we presented BrainSTR for interpretable dynamic brain-network diagnosis. BrainSTR combines Adaptive Phase Partition, critical-phase selection, structure-based topological filtering, and reference-adjusted contrastive supervision, shifting the focus from whole-sample aggregation to diagnosis-relevant spatio-temporal evidence. Experiments on ASD, BD, and MDD show consistent improvements over strong baselines. The identified critical phases and subnetworks are neurobiologically plausible, supporting the clinical interpretability of the proposed framework.

%
% ---- Bibliography ----
%
% BibTeX users should specify bibliography style 'splncs04'.
% References will then be sorted and formatted in the correct style.
%
\bibliographystyle{splncs04}
\bibliography{ref}

\end{document}